\pdfoutput=1
\newcommand{\shortversion}[1]{}
\newcommand{\longversion}[1]{#1}

\shortversion{
\documentclass[letterpaper]{article}
\usepackage{helvet}
\usepackage{courier}
\usepackage{calc}
}

\longversion{\documentclass[letter]{article}
\usepackage[totalwidth=420pt,totalheight=625pt]{geometry}
\usepackage{fixbib}

\date{}
}
 
\usepackage{times}

\shortversion{%
\frenchspacing
\usepackage{aaai}
\pdfinfo{
 /Title (Limits of Preprocessing)
 /Subject (AAAI-11)
 /Author (Stefan Szeider)
 /Keywords (Fixed-Parameter Tractability, Constraint Satisfaction, Global Constraints, Satisfiability, Bayesian Networks, Normal Logic Programs, Computational Complexity 
)}
\setcounter{secnumdepth}{0}
\usepackage[belowskip=-5pt,aboveskip=5pt]{caption} 
\leftmargin 17pt \leftmargini\leftmargin \leftmarginii 10pt
}

\usepackage{tikz}
\usetikzlibrary{decorations}

\usepackage{amsthm}
\usepackage{amsfonts}
\usepackage{amsmath}
\usepackage{url}

\makeatletter
\longversion{
\def\leftcite{(}\def\rightcite{)}
}
\makeatother

\newcommand{\citex}[1]{\citeauthor{#1}~\shortcite{#1}}
\newcommand{\citey}[1]{\citeauthor{#1},~\citeyear{#1}}

\usepackage{tikz}

\newtheorem{THE}{Theorem}

\theoremstyle{remark}

\def\hy{\hbox{-}\nobreak\hskip0pt} 

\newcommand{\SB}{\{\,}%
\newcommand{\SM}{\;{:}\;}%
\newcommand{\SE}{\,\}}%

\newcommand{\Card}[1]{|#1|}
\newcommand{\CCard}[1]{\|#1\|}
\let\phi=\varphi
\let\epsilon=\varepsilon

\newcommand{\UP}{\text{\normalfont{UP}}}
\newcommand{\Nat}{\mathbb{N}}
 
\newcommand{\AAA}{\mathcal{A}}

\newcommand{\CCC}{\mathcal{C}}

 \newcommand{\TTT}{\mathcal{T}}

\newcommand{\mtext}[1]{\text{\normalfont #1}}

\newcommand{\Prob}{\text{Pr}}
\newcommand{\dom}{\mathit{dom}}
\newcommand{\True}{\mathrm{true}}
\newcommand{\False}{\mathrm{false}}

\renewcommand{\P}{\text{\normalfont P}}
\newcommand{\NP}{\text{\normalfont NP}}
\newcommand{\coNP}{\text{\normalfont co-NP}}

\newcommand{\PR}[2]{\text{\normalfont\textsc{\bfseries #1}(#2)}}

\newcommand{\HORN}{\mtext{\sc Horn}}

\newcommand{\TWOCNF}{\mtext{2CNF}}

\makeatletter

\shortversion{
  \def\copyright@on{T}
  \setlength\titlebox{1.70in}
  \newcommand{\captionfonts}{\small}

  \long\def\@makecaption#1#2{%
    \vskip\abovecaptionskip
    \sbox\@tempboxa{{\captionfonts #1: #2}}%
    \ifdim \wd\@tempboxa >\hsize
    {\captionfonts #1: #2\par}
    \else
    \hbox to\hsize{\hfil\box\@tempboxa\hfil}%
    \fi
    \vskip\belowcaptionskip}

  \def\thm@space@setup{%
    \thm@preskip=5pt \thm@postskip=\thm@preskip}
  \renewenvironment{proof}[1][\proofname]{%
    \setlength{\topsep}{0pt}  \pushQED{\qed}%
    \normalfont 
    \trivlist  \item[\hskip\labelsep        \itshape
    #1\@addpunct{.}]\ignorespaces}{%
    \popQED\endtrivlist\@endpefalse\vspace{2pt}}

  \def\section{\@startsection {section}{1}{\z@}{-1.0ex plus
      -0.25ex minus -.1ex}{2pt plus 1pt minus .5pt}{\Large\bf\centering}}

  \def\paragraph{\@startsection{paragraph}{4}{\z@}%
    {2pt \@plus1ex \@minus.2ex}%
    {-1em}%
    {\normalfont\normalsize\bfseries}}

  \renewenvironment{proof}[1][\proofname]{\par
    \pushQED{\qed}%
    \normalfont 
    \trivlist
  \item[\hskip\labelsep
    \itshape
    #1\@addpunct{.}]\ignorespaces
  }{%
    \popQED\endtrivlist\@endpefalse
  }

  \def\itemize{%
    \ifnum \@itemdepth >\thr@@\@toodeep\else
      \advance\@itemdepth\@ne
      \edef\@itemitem{labelitem\romannumeral\the\@itemdepth}%
      \expandafter
      \list
        \csname\@itemitem\endcsname
        {\def\makelabel##1{\hss\llap{##1}}%
         \settowidth{\leftmargin}{\csname\@itemitem\endcsname}%
         \addtolength{\leftmargin}{\labelsep * \@itemdepth}%
         \setlength{\topsep}{2pt plus 1pt minus 2pt}%
         \setlength{\itemsep}{0pt}%
         \setlength{\parsep}{0pt}%
        }%
    \fi}

} 

\makeatother 

\begin{document}

\title{Limits of Preprocessing\thanks{Research funded by the ERC
    (COMPLEX REASON, Grand Reference 239962).}}

\shortversion{
 \author{Stefan Szeider\\[3pt]
 Vienna University of Technology, Vienna, Austria}
}
\longversion{
 \author{Stefan Szeider\\
 \small Vienna University of Technology, Vienna, Austria\\[-3pt]
 \small stefan@szeider.net} 
}

  \maketitle

  \begin{abstract}
    \shortversion{\begin{quote}}%
      We present a first theoretical analysis of the power of
      polynomial-time preprocessing for important combinatorial problems
      from various areas in AI. We consider problems from Constraint
      Satisfaction, Global Constraints, Satisfiability, Nonmonotonic and
      Bayesian Reasoning.  We show that, subject to a complexity
      theoretic assumption, none of the considered problems can be
      reduced by polynomial-time preprocessing to a problem kernel whose
      size is polynomial in a structural problem parameter of the input,
      such as induced width or backdoor size. Our results provide a firm
      theoretical boundary for the performance of polynomial-time
      preprocessing algorithms for the considered problems.

      \longversion{\medskip \sloppypar\noindent\emph{Keywords:}
       Fixed-Parameter Tractability,
       Constraint Satisfaction,
       Global Constraints,
       Satisfiability, Bayesian Networks, Normal Logic Programs, 
       Computational Complexity. }
\shortversion{\end{quote}}
  \end{abstract}

\section{Introduction}

Many important computational problems that arise in various areas of AI
are intractable. Nevertheless, AI research was very successful in
developing and implementing heuristic solvers that work well on
real-world instances.  An important component of virtually every solver
is a powerful polynomial-time preprocessing procedure that reduces the
problem input. For instance, preprocessing techniques for the
propositional satisfiability problem are based on Boolean Constraint
Propagation (see, e.g., \citey{EenBiere05}),
CSP solvers make use of various local consistency algorithms that filter
the domains of variables~(see, e.g., \citey{Bessiere06}); similar
preprocessing methods are used by solvers for Nonmonotonic and Bayesian
reasoning problems (see, e.g., \citey{GebserKaufmannNeumannSchaub08},
\citey{BoltGaag06}, respectively).

Until recently, no provable performance guarantees for polynomial-time
preprocessing methods have been obtained, and so preprocessing was only
subject of empirical studies.  A possible reason for the lack of
theoretical results is a certain inadequacy of the P vs NP framework for
such an analysis: if we could reduce in polynomial time an instance of
an NP-hard problem just by one bit, then we can solve the entire problem
in polynomial time by repeating the reduction step a polynomial number
of times, and $\P=\NP$ follows.

With the advent of \emph{parameterized complexity}
\cite{DowneyFellowsStege99}, a new theoretical framework became
available that provides suitable tools to analyze the power of
preprocessing. Parameterized complexity considers a problem in a
two-dimensional setting, where in addition to the input size~$n$, a
\emph{problem parameter~$k$} is taken into consideration.  This
parameter can encode a structural aspect of the problem instance. A
problem is called \emph{fixed-parameter tractable} (FPT) if it can be
solved in time $f(k)p(n)$ where $f$ is a function of the parameter $k$
and $p$ is a polynomial of the input size $n$. Thus, for FPT problems,
the combinatorial explosion can be confined to the parameter and is
independent of the input size.  It is known that a problem is
fixed-parameter tractable if and only if every problem input can be
reduced by polynomial-time preprocessing to an equivalent input whose
size is bounded by a function of the parameter
\cite{DowneyFellowsStege99}. The reduced instance is called the
\emph{problem kernel}, the preprocessing is called 
\emph{kernelization}. The power of polynomial-time preprocessing can now
be benchmarked in terms of the size of the kernel.  Once a small kernel
is obtained, we can apply any method of choice to solve the kernel:
brute-force search, heuristics, approximation,
etc.~\cite{GuoNiedermeier07}.  Because of this flexibility a small
kernel is generally preferable to a less flexible branching-based
fixed-parameter algorithm. Thus, small kernels provide an additional
value that goes beyond bare fixed-parameter tractability.

In general the size of the kernel is exponential in the parameter, but
many important $\NP$-hard optimization problems such as Minimum Vertex
Cover, parameterized by solution size, admit \emph{polynomial kernels},
see, e.g., \cite{BodlaenderDowneyFellowsHermelin09} for references.

In previous research several NP-hard AI problems have been shown to be
fixed-parameter tractable. We list some important examples from various
areas:

\begin{itemize}
\item Constraint satisfaction problems (CSP) over a fixed universe of
  values, parameterized by the induced
  width~\cite{GottlobScarcelloSideri02}.
\item Consistency and generalized arc consistency for intractable global
  constraints, parameterized by the cardinalities of certain sets of
  values \cite{BessiereEtal08}.
\item Propositional satisfiability (SAT), parameterized by the size of
  backdoors~\cite{NishimuraRagdeSzeider04-informal}.
\item Positive inference in Bayesian networks with variables of bounded
  domain size, parameterized by size of loop
  cutsets~\cite{Pearl88,BidyukDechter07}.
\item Nonmonotonic reasoning with normal logic programs, parameterized by
  feedback width~\cite{GottlobScarcelloSideri02}.
\end{itemize}
However, only exponential kernels are known for these fundamental AI
problems.  
Can we hope for polynomial kernels?

\paragraph{Results}
Our results are throughout negative. We provide strong theoretical
evidence that none of the above fixed-parameter tractable AI problems
admits a polynomial kernel. More specifically, we show that a polynomial
kernel for any of these problems causes a collapse of the Polynomial
Hierarchy to its third level, which is considered highly unlikely by
complexity theorists.

Our results are general: The kernel lower bounds are not limited to a
particular preprocessing technique but apply to any clever technique
that could be conceived in future research. Hence the results contribute
to the foundations of~AI.

Our results suggest the investigation of alternative approaches to
polynomial-time preprocessing; for instance, preprocessing that produces
in polynomial time a Boolean combination of polynomially sized kernels
instead of one single kernel.

\section{Formal Background}

\newcommand{\PP}{\normalfont \textbf{P}}
\newcommand{\QQ}{\normalfont \textbf{Q}}

A \emph{parameterized problem} $\PP$ is a subset of $\Sigma^* \times
\Nat$ for some finite alphabet $\Sigma$. For a problem instance $(x,k)
\in \Sigma^* \times \Nat$ we call $x$ the main part and $k$ the
parameter. We assume the parameter is represented in unary.  For the
parameterized problems considered in this paper, the parameter is a
function of the main part, i.e., $k=\pi(x)$ for a function $\pi$. We
then denote the problem as $\PP(\pi)$, e.g.,  $\PR{$U$-CSP}{width}$
denotes the problem \textbf{$U$-CSP} parameterized by the width of the
given tree decomposition.

A parameterized problem $\PP$ is \emph{fixed-parameter tractable} if
there exists an algorithm that solves any input $(x,k) \in \Sigma^*
\times \Nat$ in time $O(f(k) \cdot p(\Card{x})$ where $f$ is an
arbitrary computable function of $k$ and $p$ is a polynomial in~$n$.

A \emph{kernelization} for a parameterized problem $\PP \subseteq
\Sigma^* \times \Nat$ is an algorithm that, given $(x, k) \in \Sigma^*
\times \Nat$, outputs in time polynomial in $\Card{x}+k$ a pair $(x',
k') \in \Sigma^* \times \Nat$ such that (i)~$(x,k) \in \PP$ if and only
if $(x',k') \in \PP$ and (ii)~$\Card{x'}+k' \leq g(k)$, where $g$ is an
arbitrary computable function, called the \emph{size} of the kernel.  In
particular, for constant $k$ the kernel has constant size $g(k)$.  If
$g$ is a polynomial then we say that $\PP$ admits a \emph{polynomial
  kernel}.

Every fixed-parameter tractable problem admits a kernel. This can be
seen by the following argument due to \citex{DowneyFellowsStege99}.
Assume we can decide instances $(x,k)$ of problem $\PP$ in time
$f(k)\Card{n}^{O(1)}$. We kernelize an instance $(x,k)$ as follows. If
$\Card{x}\leq f(k)$ then we already have a kernel of size $f(k)$.
Otherwise, if $\Card{x}> f(k)$, then $f(k)\Card{x}^{O(1)}\leq
\Card{x}^{O(1)}$ is a polynomial; hence we can decide the instance in
polynomial time and replace it with a small decision-equivalent instance
$(x',k')$. Thus we always have a kernel of size at most $f(k)$.
However, $f(k)$ is super-polynomial for NP-hard problems (unless
$\P=\NP$), hence this generic construction is not providing polynomial
kernels.

We understand \emph{preprocessing} for an NP-hard problem as a
\emph{polynomial-time} procedure that transforms an instance of the
problem to a (possible smaller) solution-equivalent instance of the same
problem.  Kernelization is such a preprocessing with a \emph{performance
  guarantee}, i.e., we are guaranteed that the preprocessing yields a
kernel whose size is bounded in terms of the parameter of the given
problem instance. In the literature also different forms of
preprocessing have been considered. An important one is \emph{knowledge
  compilation}, a two-phases approach to reasoning problems where in a
first phase a given knowledge base is (possibly in exponential time)
preprocessed (``compiled''), such that in a second phase various queries
can be answered in polynomial time~\cite{CadoliDoniniLiberatore02}.

\section{Tools for Kernel Lower Bounds}

In the sequel we will use recently developed tools to obtain kernel
lower bounds. Our kernel lower bounds are subject to the widely believed
complexity theoretic assumption $\NP \not\subseteq \coNP/\text{poly}$
(or equivalently, $\text{PH}\neq \Sigma_p^3$). In other words, the tools
allow us to show that a parameterized problem does not admit a
polynomial kernel unless the Polynomial Hierarchy collapses to its third
level (see, e.g., \citey{Papadimitriou94}).

A \emph{composition algorithm} for a parameterized problem $\PP
\subseteq \Sigma^* \times \Nat$ is an algorithm that receives as input a
sequence $(x_1, k),\dots,(x_t, k) \in \Sigma^* \times \Nat$, uses time
polynomial in $\sum_{i=1}^t \Card{x_i}+k$, and outputs $(y,k') \in
\Sigma^* \times \Nat$ with (i)~$(y,k')\in \PP$ if and only if $(x_i,k)
\in \PP$ for some $1 \leq i \leq t$, and (ii)~$k'$ is polynomial in~$k$.
A parameterized problem is \emph{compositional} if it has a composition
algorithm.  With each parameterized problem $\PP\subseteq \Sigma^*
\times \Nat$ we associate a classical problem 
\shortversion{$\UP[\PP]=\SB x\#1^k \SM (x,k)\in P \SE$ }
\longversion{\[\UP[\PP]=\SB x\#1^k \SM (x,k)\in P \SE\] }
where $1$ denotes an arbitrary symbol from $\Sigma$ and
$\#$ is a new symbol not in $\Sigma$.  We call $\UP[\PP]$ the
\emph{unparameterized version} of~$\PP$.

The following result is the basis for our kernel lower bounds.

\begin{THE}[\citey{BodlaenderDowneyFellowsHermelin09},
  \citey{FortnowSanthanam08}]\label{the:comp}
  Let $\PP$ be a parameterized problem whose unparameterized version is
  $\NP$-complete. If $\PP$ is compositional, then it does not admit a
  polynomial kernel unless $\NP \subseteq \coNP/\text{poly}$, i.e., the
  Polynomial Hierarchy collapses.
\end{THE}

Let $\PP,\QQ\subseteq \Sigma^* \times \Nat$ be parameterized
problems. We say that $\PP$ is \emph{polynomial parameter reducible} to
$\QQ$ if there exists a polynomial time computable function $K: \Sigma^*
\times \Nat \rightarrow \Sigma^* \times \Nat$ and a polynomial $p$, such
that for all $(x,k) \in \Sigma^* \times \Nat$ we have (i)~$(x,k) \in
\PP$ if and only if $K(x,k) =(x',k')  \in \QQ$, and (ii)~$k'\leq
p(k)$. The function $K$ is called a \emph{polynomial parameter
  transformation}.

The following theorem allows us to transform kernel lower bounds from
one problem to another.
\begin{THE}[\citey{BodlaenderThomasseYeo09}]\label{the:trans}
  Let $\PP$ and $\QQ$ be  parameterized problems such that $\UP[\PP]$
  is $\NP$-complete, $\UP[\QQ]$ is in $\NP$, and there is a polynomial
  parameter transformation from $\PP$ to~$\QQ$. If $\QQ$ has a
  polynomial kernel, then $\PP$ has a polynomial kernel.
\end{THE}

\section{Constraint Networks}

\emph{Constraint networks} have proven successful in modeling everyday
cognitive tasks such as vision, language comprehension, default
reasoning, and abduction, as well as in applications such as scheduling,
design, diagnosis, and temporal and spatial reasoning \cite{Dechter10}.
A \emph{constraint network} is a triple $I = (V, U, \CCC)$ where $V$ is
a finite set of variables, $U$ is a finite universe of values, and
$\CCC=\{C_1,\dots,C_m\}$ is set of constraints. Each constraint $C_i$ is
a pair $(S_i,R_i)$ where $S_i$ is a list of variables of length $r_i$
called the \emph{constraint scope}, and $R_i$ is an $r_i$-ary relation
over $U$, called the \emph{constraint relation}. The tuples of $R_i$
indicate the allowed combinations of simultaneous values for the
variables $S_i$. A \emph{solution} is a mapping $\tau :V \rightarrow U$
such that for each $1\leq i \leq m$ and $S_i=(x_1,\dots,x_{r_i})$, we
have $(\tau(x_1),\dots,\tau(x_{r_i}))\in R_i$.  A constraint network is
\emph{satisfiable} if it has a solution.

With a constraint network $I = (V, U, \CCC)$ we associate its
\emph{constraint graph} $G=(V,E)$ where $E$ contains an edge between two
variables if and only if they occur together in the scope of a
constraint.  A \emph{width $w$ tree decomposition} of a graph $G$ is a
pair $(T,\lambda)$ where $T$ is a tree and $\lambda$ is a labeling of
the nodes of $T$ with sets of vertices of $G$ such that the following
properties are satisfied: (i)~every vertex of $G$ belongs to
$\lambda(p)$ for some node $p$ of~$T$; (ii)~every edge of $G$ is is
contained in $\lambda(p)$ for some node $p$ of~$T$; (iii)~For each
vertex $v$ of $G$ the set of all tree nodes $p$ with $v\in \lambda(p)$
induces a connected subtree of $T$; (iv)~$\Card{\lambda(p)}-1\leq w$
holds for all tree nodes~$p$.  The \emph{treewidth} of $G$ is the
smallest $w$ such that $G$ has a width $w$ tree decomposition.  The
\emph{induced width} of a constraint network is the treewidth of its
constraint graph \cite{DechterPearl89}. We note in passing that the
problem of finding a tree decomposition of width $w$ is $\NP$-hard but
fixed-parameter tractable in~$w$.

Let $U$ be a fixed universe containing at least two elements.  We
consider the following parameterized version of the constraint
satisfaction problem (CSP).  \longversion{
\begin{quote}
  $\PR{$U$-CSP}{width}$

  \emph{Instance:} A constraint network $I=(V,U,\CCC)$ and a width $w$ tree
  decomposition of the constraint graph of $I$.

  \emph{Parameter:} The integer $w$.

  \emph{Question:} Is $I$ satisfiable?
\end{quote}} \shortversion{\par $\PR{$U$-CSP}{width}$: the instance is
constraint network $I=(V,U,\CCC)$ and a width $w$ tree decomposition of
the constraint graph of $I$.  $w$ is the parameter. The question is
whether is $I$ satisfiable.\par}
It is well known that $\PR{$U$-CSP}{width}$ is fixed-parameter tractable
over any fixed universe
$U$~\cite{DechterPearl89,GottlobScarcelloSideri02} (for generalizations
see \citey{SamerSzeider10a}).  We contrast this
classical result and show that it is unlikely that $\PR{$U$-CSP}{width}$
admits a polynomial kernel, even in the simplest case where $U=\{0,1\}$.
\begin{THE}\label{the:csp}
  $\PR{$\{0,1\}$-CSP}{width}$ does not admit a polynomial kernel
  unless the Polynomial Hierarchy collapses.
\end{THE}
\begin{proof}
  We show that $\PR{$\{0,1\}$-CSP}{width}$ is compositional.  Let
  $(I_i,T_i)$, $1\leq i \leq t$, be a given sequence of instances of
  $\PR{$\{0,1\}$-CSP}{width}$ where $I_i=(V_i,U_i,\CCC_i)$ is a
  constraint network and $T_i$ is a width $w$ tree decomposition of the
  constraint graph of $I_i$.  We may assume, w.l.o.g., that $V_i \cap
  V_j =\emptyset$ for $1\leq i < j \leq t$ (otherwise we can simply
  change the names of variables).  We form a new constraint network
  $I=(V,\{0,1\},\CCC)$ as follows.  We put $V=\bigcup_{i=1}^t V_i \cup
  \{a_1,\dots,a_t,b_0,\dots,b_t\}$ where $a_i,b_i$ are new variables. We
  define the set $\CCC$ of constraints in three groups.

  (1)~For each $1\leq i \leq t$ and each constraint
  $C=((x_1,\dots,x_r),R)\in \CCC_i$ we add to $\CCC$ a new constraint
  $C'=((x_1,\dots,x_r,a_i),R'))$ where $R'=\SB (u_1,\dots,u_r,0) \SM
  (u_1,\dots,u_r)\in R \SE \cup \{(1,\dots,1)\}$. 

  (2) We add $t$ ternary constraints $C_1^*,\dots,C_t^*$ where
  $C_i^*=((b_{i-1},b_{i},a_i),R^*)$ and $R^*=\{(0,0,1)$, $(0,1,0)$,
  $(1,1,1)\}$.

  (3) Finally, we add two unary constraints $C^0=((b_0),(0))$ and
  $C^1=((b_t),(1))$ which force the values of $b_0$ and $b_t$ to $0$ and
  $1$, respectively.

  Let $G, G_i$ be the constraint graphs of $I$ and $I_i$, respectively.
\longversion{Fig.~\ref{fig:CG} shows an illustration of~$G$ for $t=4$.
\begin{figure}[tbp]
  \centering
 \begin{tikzpicture}
    \small
    \draw
    (0,0)   node (b0) {$b_0$}
    (1,0)   node (b1) {$b_1$}
    (2,0)   node (b2) {$b_2$}
    (3,0)   node (b3) {$b_3$}
    (4,0)   node (b4) {$b_4$}

    (0.5,.8)   node (a1) {$a_1$}
    (1.5,.8)   node (a2) {$a_2$}
    (2.5,.8)   node (a3) {$a_3$}
    (3.5,.8)   node (a4) {$a_4$}

    (0.5-.3,1.8)   coordinate (x1) {}
    (0.5+.3,1.8)   coordinate (y1) {}
    (0.5,1.8)   node (z1) {$\dots$}
    (0.5,2.1)   node (V1) {$V_1$}

    (1.5-.3,1.8)   coordinate (x2) {}
    (1.5+.3,1.8)   coordinate (y2) {}
    (1.5,1.8)   node (z2) {$\dots$}
    (1.5,2.1)   node (V2) {$V_2$}

    (2.5-.3,1.8)   coordinate (x3) {}
    (2.5+.3,1.8)   coordinate (y3) {}
    (2.5,1.8)   node (z3) {$\dots$}
    (2.5,2.1)   node (V3) {$V_3$}

    (3.5-.3,1.8)   coordinate (x4) {}
    (3.5+.3,1.8)   coordinate (y4) {}
    (3.5,1.8)   node (z4) {$\dots$}
    (3.5,2.1)   node (V4) {$V_4$}

    ;

    \draw (b0)--(b1)--(b2)--(b3)--(b4)
    (b0)--(a1)--(b1)--(a2)--(b2)--(a3)--(b3)--(a4)--(b4)
    (a1)--(x1) (a1)--(y1)
    (a2)--(x2) (a2)--(y2)
    (a3)--(x3) (a3)--(y3)
    (a4)--(x4) (a4)--(y4)

    ;

 \end{tikzpicture}%
 \caption{Constraint graph $G$.} \label{fig:CG}
 \end{figure}
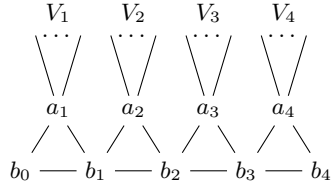}
We observe that $a_1,\dots,a_t$ are cut vertices of~$G$. Removing these
vertices separates $G$ into independent parts $P,G_1',\dots,G_t'$ where
$P$ is the path $b_0,b_1,\dots,b_t$, and $G_i'$ is isomorphic to $G_i$.
By standard techniques (see, e.g., \citey{Kloks94}), we can put the given
width $w$ tree decompositions $T_1,\dots,T_t$ of $G_1',\dots,G_t'$ and
the trivial width~1 tree decomposition of $P$ together to a width $w+1$
tree decomposition $T$ of~$G$. Clearly $(I,T)$ can be obtained from
$(I_i,T_i)$, $1\leq i \leq t$, in polynomial time. 

\shortversion{It is not difficult to see that $I$ is satisfiable if and
  only if at least one of the $I_i$ is satisfiable.}
\longversion{
We claim that $I$ is satisfiable if and only if at least one of the
$I_i$ is satisfiable.  
This claim can be verified by means of the
following observations: The constraints in groups (2) and (3) provide
that for any satisfying assignment there will be some $0\leq i \leq t-1$
such that $b_0,\dots,b_i$ are all set to 0 and $b_{i+1},\dots,b_t$ are
all set to $1$; consequently $a_i$ is set to $0$ and all $a_j$ for
$j\neq i$ are set to~1. The constraints in group (1) provide that if we
set $a_i$ to $0$, then we obtain from $C'$ the original constraint $C$;
if we set $a_i$ to $1$ then we obtain a constraint that can be satisfied
by setting all remaining variables to~$1$. We conclude that
$\PR{$\{0,1\}$-CSP}{width}$ is compositional.}

\longversion{\sloppypar In order to apply Theorem~\ref{the:comp}, it
  remains to establish that the unparameterized version of
  $\PR{$\{0,1\}$-CSP}{width}$ is $\NP$-complete. Deciding whether a
  constraint network~$I$ over the universe $\{0,1\}$ is satisfiable is
  well-known to be $\NP$-complete (say by reducing 3-SAT). To a
  constraint network $I$ on $n$ variables we can always add a trivial
  width $w=n-1$ tree decomposition of its constraint graph (taking a
  single tree node $t$ where $\lambda(t)$ contains all variables of
  $I$). Hence $\UP[\PR{$\{0,1\}$-CSP}{width}]$ is $\NP$-complete.}

\shortversion{In order to apply Theorem~\ref{the:comp}, it remains to
  observe that $\UP[\PR{$\{0,1\}$-CSP}{width}]$ is $\NP$-complete.}
\end{proof}

\longversion{\enlargethispage*{5mm}}
\section{Satisfiability}
The \emph{propositional satisfiability problem} (SAT) was the first problem
shown to be NP-hard \cite{Cook71}. Despite its hardness, SAT solvers are
increasingly leaving their mark as a general-purpose tool in areas as
diverse as software and hardware verification, automatic test pattern
generation, planning, scheduling, and even challenging problems from
algebra \cite{GomesKautzSabharwalSelman08}.  SAT solvers are capable of
exploiting the hidden structure present in real-world problem instances.
The concept of \emph{backdoors}, introduced by
\citex{WilliamsGomesSelman03a} provides a means for making the vague
notion of a hidden structure explicit.  Backdoors are defined with
respect to a ``sub-solver'' which is a polynomial-time algorithm that
correctly decides the satisfiability for a class $\CCC$ of CNF
formulas. More specifically, \citex{GomesKautzSabharwalSelman08} define
a \emph{sub-solver} to be an algorithm $\AAA$ that takes as input a CNF
formula $F$ and has the following properties: (i)~\emph{Trichotomy}:
$\AAA$ either rejects the input $F$, or determines $F$ correctly as
unsatisfiable or satisfiable; (ii)~\emph{Efficiency}: $\AAA$ runs in
polynomial time; (iii)~\emph{Trivial Solvability}: $\AAA$ can determine
if $F$ is trivially satisfiable (has no clauses) or trivially
unsatisfiable (contains only the empty clause); (iv.)
\emph{Self-Reducibility}: if $\AAA$ determines $F$, then for any
variable $x$ and value $\epsilon\in\{0,1\}$, $\AAA$ determines
$F[x=\epsilon]$.  $F[\tau]$ denotes the formula obtained from $F$ by
applying the partial assignment $\tau$, i.e., satisfied clauses are
removed and false literals are removed from the remaining clauses.

\longversion{\enlargethispage*{5mm}}
We identify a sub-solver $\AAA$ with the class $\CCC_\AAA$ of CNF
formulas whose satisfiability can be determined by $\AAA$.  A
\emph{strong} \emph{$\AAA$-backdoor set} (or \emph{$\AAA$-backdoor}, for
short) of a CNF formula $F$ is a set $B$ of variables such that for each
possible truth assignment $\tau$ to the variables in $B$, the
satisfiability of $F[\tau]$ can be determined by sub-solver $\AAA$ in
time $O(n^c)$. Hence, if we know an $\AAA$-backdoor of size $k$, we can
decide the satisfiability of $F$ by running $\AAA$ on $2^k$ instances
$F[\tau]$, yielding a time bound of $O(2^k n^c)$. Hence SAT decision is
fixed-parameter tractable in the backdoor size~$k$ for any sub-solver
$\AAA$.  Hence the following problem is clearly fixed-parameter
tractable for any sub-solver~$\AAA$.
\longversion{
\begin{quote}
  $\PR{SAT}{$\AAA$-backdoor}$

  \emph{Instance:} A CNF formula $F$, and an $\AAA$-backdoor $B$ of $F$
  of size $k$. 

 \emph{Parameter:} The integer $k$. 

 \emph{Question:} Is
  $F$ satisfiable?
\end{quote}}
\shortversion{
 \par$\PR{SAT}{$\AAA$-backdoor}$: the instance is a CNF formula $F$, and an
 $\AAA$-backdoor $B$ of $F$  of size $k$. The parameter is $k$, the
 question is whether $F$ is satisfiable.\par}
We are concerned with the question of whether instead of trying all
$2^k$ possible partial assignments we can reduce the instance to a
polynomial kernel.  We will establish a very general result that applies
to all possible sub-solvers.
\begin{THE}\label{the:backdoor}
  $\PR{SAT}{$\AAA$-backdoor}$ does not admit a polynomial kernel for any
  sub-solver $\AAA$ unless the Polynomial Hierarchy collapses.
\end{THE}
\begin{proof}
  We will devise polynomial parameter transformations from the following
  parameterized problem which is known to be compositional
  \cite{FortnowSanthanam08} and therefore  unlikely to admit a
  polynomial kernel.
  \longversion{
  \begin{quote}
    $\PR{SAT}{vars}$

    \emph{Instance:} A propositional formula $F$ in CNF on $n$
    variables.

    \emph{Parameter:} The number $n$ of variables.

    \emph{Question:} Is $F$ satisfiable?
  \end{quote}}
\shortversion{
\par$\PR{SAT}{vars}$: the instance is a CNF formula $F$ on $n$
    variables. The parameter is $n$, the question is whether $F$ is
    satisfiable. \par}
  Let $F$ be a CNF formula and $V$ the set of all variables of~$F$. Due
  to property~(ii) of a sub-solver, $V$ is an $\AAA$-backdoor set
  for any $\AAA$.  Hence, by mapping $(F,n)$ (as an instance of
  $\PR{SAT}{vars}$) to $(F,V)$ (as an instance of
  $\PR{SAT}{$\AAA$-backdoor}$) provides a (trivial) polynomial parameter
  transformation from $\PR{SAT}{vars}$ to
  $\PR{SAT}{$\AAA$-backdoor}$.  Since the unparameterized versions of
  both problems are clearly NP-complete, the result follows by
  Theorem~\ref{the:trans}.
\end{proof}

Let $\PR{3SAT}{$\pi$}$ (where $\pi$ is an arbitrary parameterization)
denote the problem $\PR{SAT}{$\pi$}$ restricted to 3CNF formula, i.e.,
to CNF formulas where each clauses contains at most three literals. In
contrast to $\PR{SAT}{vars}$, the parameterized problem
$\PR{3SAT}{vars}$ has a trivial polynomial kernel: if we remove
duplicate clauses, then any 3CNF formula on $n$ variables contains at
most $O(n^3)$ clauses, and so is a polynomial kernel.  Hence the easy
proof of Theorem~\ref{the:backdoor} does not carry over to
$\PR{3SAT}{$\AAA$-backdoor}$. We therefore consider the cases
$\PR{3SAT}{$\HORN$-backdoor}$ and $\PR{3SAT}{$\TWOCNF$-backdoor}$
separately, these cases are important since the detection of $\HORN$ and
$\TWOCNF$-backdoors is fixed-parameter tractable
\cite{NishimuraRagdeSzeider04-informal}.

\begin{THE}
  Neither $\PR{3SAT}{$\HORN$-backdoor}$ nor
  $\PR{3SAT}{$\TWOCNF$-backdoor}$ admit a polynomial kernel unless the
  Polynomial Hierarchy collapses.
\end{THE}
\longversion{\begin{proof}
  Let $\CCC\in \{\HORN,\TWOCNF\}$.  We show that
  $\PR{3SAT}{$\CCC$-backdoor}$ is compositional.  Let $(F_i,B_i)$,
  $1\leq i \leq t$, be a given sequence of instances of
  $\PR{3SAT}{$\CCC$-backdoor}$ where $F_i$ is a 3CNF formula and $B_i$
  is a $\CCC$-backdoor set of $F_i$ of size~$k$.  We distinguish two
  cases. 

  Case 1: $t>2^k$.  Let $\CCard{F_i}:=\sum_{C\in F_i} \Card{C}$ and
  $n:=\max_{i=1}^t \CCard{F_i}$.  Whether $F_i$ is satisfiable or not
  can be decided in time $O(2^kn)$ since the satisfiability of a Horn or
  2CNF formula can be decided in linear time. We can check whether at
  least one of the formulas $F_1,\dots,F_t$ is satisfiable in time
  $O(t2^k n)\leq O(t^2n)$ which is polynomial in $t+n$.  If some $F_i$
  is satisfiable, we output $(F_i,B_i)$; otherwise we output $(F_1,B_1)$
  ($F_1$ is unsatisfiable). Hence we have a composition algorithm.

  Case 2: $t\leq 2^k$. This case is more involved.  We construct a new
  instance $(F,B)$ of $\PR{3SAT}{$\CCC$-backdoor}$ as follows. 

  Let $s=\lceil \log_2 t \rceil$.  Since $t\leq 2^k$, $s\leq k$ follows.

  Let $V_i$ denote the set of variables of $F_i$. We may assume,
  w.l.o.g., that $B_1=\dots=B_t$ and that $V_i\cap V_j =B_1$ for all
  $1\leq i < j \leq t$ since otherwise we can change names of variable
  accordingly.  In a first step we obtain from every $F_i$ a CNF formula
  $F_i'$ as follows. For each variable $x\in V_i\setminus B_1$ we take
  two new variables $x_0$ and $x_1$. We replace each positive occurrence
  of a variable $x\in V_i\setminus B_1$ in $F_i$ with the literal $x_0$
  and each negative occurrence of $x$ with the literal $\neg x_s$.  We add
  all clauses of the form $(\neg x_{j-1} \vee x_j)$ for $1\leq j \leq s$;
  we call these clauses ``\emph{connection clauses}.'' Let $F_i'$ be the
  formula obtained from $F_i$ in this way. We observe that $F_i'$ and
  $F_i$ are SAT-equivalent, since the connection clauses form an
  implication chain. Since the connection clauses are both
  Horn and 2CNF, $B_1$ is also a $\CCC$-backdoor of $F_i'$.
  
  We take a set $Y=\{y_1,\dots,y_s\}$ of new variables. Let
  $C_1,\dots,C_{2^s}$ be the sequence of all $2^s$ possible clauses
  (modulo permutation of literals within a clause) containing exactly
  $s$ literals over the variables in~$Y$.  Consequently we can write
  $C_i$ as $(\ell^i_1 \vee \dots \vee \ell^i_s)$ where $\ell_i^j\in
  \{y_i,\neg y_i\}$.
  \par
  For $1\leq i \leq t$ we add to each connection clause $(\neg x_{j-1} \vee
  x_j)$
  of $F_i'$ the literal $\ell^i_j\in C_i$. Let $F_i''$ denote the 3CNF
  formula obtained from $F_i'$ this way.
   
  For $t< i \leq 2^s$ we define 3CNF formulas $F_i''$ as follows.  If
  $s\leq 3$ then $F_i''$ consists just of the clause~$C_i$.  If $s>3$
  then we take new variables $z^i_2,\dots,z^i_{s-2}$ and let $F_i''$
  consist of the clauses $(\ell^i_1 \vee \ell^i_2 \vee \neg z^i_2)$,
  $(\ell^i_3 \vee z^i_2 \vee \neg z^i_3), \dots, (\ell^i_{s-2} \vee 
  z^i_{s-3} \vee \neg z^i_{s-2})$, $(\ell^i_{s-1} \vee \ell^i_{s} \vee 
  z^i_{s-2})$.  Finally, we let $F$ be the 3CNF formula containing all
  the clauses from $F_1'',\dots,F_{2^s}''$.  Any assignment $\tau$ to
  $Y\cup B_1$ that satisfies $C_i$  can be extended
  to an assignment that satisfies $F_i''$ since such assignment
  satisfies at least one connection clause $(x_{j-1} \vee x_j \vee
  \ell^i_j)$ and so the chain of implications from from $x_o$ to $x_s$
  is broken. 

  \enlargethispage*{8mm}
  It is not difficult to verify the following two claims. (i)~$F$ is
  satisfiable if and only if at least one of the formulas $F_i$ is
  satisfiable.  (ii)~$B=Y \cup B_1$ is a $\CCC$-backdoor of $F$.  Hence
  we have also a composition algorithm in Case~2, and thus
  $\PR{3SAT}{$\CCC$-backdoor}$ is compositional. Clearly
  $\UP[\PR{3SAT}{$\CCC$-backdoor}]$ is $\NP$\hy complete, hence the
  result follows from~Theorem~\ref{the:comp}.
\end{proof}} \shortversion{\begin{proof}(Sketch; for more details see
  \url{http://arxiv.org/abs/YYYY.XXXX}.) Let $\CCC\in \{\HORN,\TWOCNF\}$.
  We show that $\PR{3SAT}{$\CCC$-backdoor}$ is compositional.  Let
  $(F_i,B_i)$, $1\leq i \leq t$, be a given sequence of instances of
  $\PR{3SAT}{$\CCC$-backdoor}$ where $F_i$ is a 3CNF formula and $B_i$
  is a $\CCC$-backdoor set of $F_i$ of size~$k$.

  If $t>2^k$ then we can determine whether some $F_i$ is satisfiable in
  time $O(t2^k n)\leq O(t^2n)$ which is polynomial in $t+n$.  If the
  answer is yes, then we output $(F_i,B_i)$, otherwise we output
  $(F_1,B_1)$.  It remains to consider the case where $t\leq 2^k$. For
  simplicity, assume $t= 2^k$.  Let $V_i$ denote the set of variables of
  $F_i$. We may assume, w.l.o.g., that $B_1=\dots=B_t$ and that $V_i\cap
  V_j =B_1$ for all $1\leq i < j \leq t$ since otherwise we can change
  names of variable accordingly.  We take a set $Y=\{y_1,\dots,y_s\}$ of
  new variables. From each $F_i$ we construct a formula $F_i'$ such that
  for the $i$-th truth assignment $\tau_i$ to $Y$, $F_i'[\tau_i]$ and
  $F_i$ are decision-equivalent, and $F_j'[\tau_i]$ is trivially
  satisfiable for $j\neq i$. This can be done such that (i)~$F$ is
  satisfiable if and only if at least one of the formulas $F_i$ is
  satisfiable and (ii)~$B=Y \cup B_1$ is a $\CCC$-backdoor of $F$.
  Hence we have a composition algorithm for
  $\PR{3SAT}{$\CCC$-backdoor}$. Since $\UP[\PR{3SAT}{$\CCC$-backdoor}]$
  is clearly $\NP$\hy complete, the result follows
  from~Theorem~\ref{the:comp}.
\end{proof}}

\section{Global Constraints}

The success of today's constraint solvers relies heavily on efficient
algorithms for special purpose \emph{global
  constraints}~\cite{HoeveKatriel06}.  A global constraint specifies a
pattern that frequently occurs in real-world problems, for instance, it
is often required that variables must all take different values (e.g.,
activities requiring the same resource must all be assigned different
times). The \textsc{AllDifferent} global constraint efficiently encodes
this requirement.  

More formally, a global constraint is defined for a set $S$ of
variables, each variable $x\in S$ ranges over a finite domain $\dom(x)$
of values. An \emph{instantiation} is an assignment $\alpha$ such that
$\alpha(x)\in \dom(x)$ for each $x\in S$.  A global constraint defines
which instantiations are legal and which are not.  A
global constraint is \emph{consistent} if it has at least one legal
instantiation, and it is \emph{domain consistent} (or hyper arc
consistent) if for each variable $x\in S$ and each value $d\in \dom(x)$
there is a legal instantiation $\alpha$ with $\alpha(x)=d$. For all
global constraints considered in this paper, domain consistency can be
reduced to a quadratic number of consistency checks, hence we will focus
on consistency. We assume that the size of a representation of a global
constraint is polynomial in $\sum_{x\in S} \Card{\dom(x)}$.

For several important types $\TTT$ of global constraints, the problem of
deciding whether a constraint of type $\TTT$ is consistent (in symbols
$\TTT$-\textbf{Cons}) is $\NP$-hard. Examples for such intractable
types of constraints are \textsc{NValue}, \textsc{Disjoint}, and
\textsc{Uses}~\cite{BessiereEtAl04}.  An \textsc{NValue} constraint over
a set $X$ of variables requires from a legal instantiation $\alpha$ that
$\Card{ \SB \alpha(x) \SM x\in X \SE}=N$; \textsc{AllDifferent} is the
special case where $N=\Card{X}$. The global constraints
\textsc{Disjoint} and \textsc{Uses} are specified by two sets of
variables $X,Y$; \textsc{Disjoint} requires that $\alpha(x)\neq
\alpha(y)$ for each pair $x\in X$ and $y\in Y$; \textsc{Uses} requires
that for each $x\in X$ there is some $y \in Y$ such that
$\alpha(x)=\alpha(y)$.  For a set $X$ of variables we write
$\dom(X)=\bigcup_{x\in X} \dom(x)$.

\citex{BessiereEtal08} considered $dx=\Card{\dom(X)}$ as
parameter for \textsc{NValue}, $dxy=\Card{\dom(X) \cap \dom(Y)}$ as
parameter for \textsc{Disjoint}, and $dy=\Card{\dom(Y)}$ as parameter
for \textsc{Uses}.  They showed that consistency checking is
fixed-parameter tractable for the constraints under the respective
parameterizations, i.e., the problems \textsc{NValue}\hy
$\PR{Cons}{$dx$}$, \textsc{Disjoint}\hy $\PR{Cons}{$dxy$}$, and
\textsc{Uses}\hy $\PR{Cons}{$dy$}$ are fixed-parameter tractable.  We
show that it is unlikely that their results can be improved in terms of
polynomial kernels.

\begin{THE} The problems \textsc{NValue}\hy $\PR{Cons}{$dx$}$,
  \textsc{Disjoint}\hy $\PR{Cons}{$dxy$}$, \textsc{Uses}\hy
  $\PR{Cons}{$dy$}$ do not admit  polynomial kernels unless the
  Polynomial Hierarchy collapses.
\end{THE}
\begin{proof} We devise a polynomial parameter reduction from
  $\PR{SAT}{vars}$.  We use a construction of~\citex{BessiereEtAl04}.
  Let $F= \{C_1,\dots,C_m\}$ be a CNF formula over variables
  $x_1,\dots,x_n$.  We consider the clauses and variables of $F$ as the
  variables of a global constraint with domains $\dom(x_i)=\{-i,i\}$,
  and $\dom(C_j)=\SB i \SM x_i\in C_j \SE \cup \SB -i \SM \neg x_i\in
  C_j \SE$. Now $F$ can be encoded as an \textsc{NValue} constraint with
  $X=\{x_1,\dots,x_n,C_1,\dots,C_m\}$ and $N=n$ (clearly $F$ is
  satisfiable if and only if the constraint is consistent).  Since
  $dx=2n$ we have a polynomial parameter reduction from $\PR{SAT}{vars}$
  to \textsc{NValue}\hy $\PR{Cons}{$dx$}$.  Similarly, as observed by
  \citex{BessiereHebrardHnichKiziltanWalsh09}, $F$ can be encoded as a
  \textsc{Disjoint} constraint with $X=\{x_1,\dots,x_n\}$ and
  $Y=\{C_1,\dots,C_m\}$ ($dxy\leq 2n$), or as a \textsc{Uses} constraint
  with $X=\{C_1,\dots,C_m\}$ and $Y=\{x_1,\dots,x_n\}$ ($dy=2n$).
  Since the unparameterized problems are clearly NP-complete, the result
  follows by Theorem~\ref{the:trans}.
\end{proof}
\enlargethispage*{5mm}
\noindent Further results on kernels for global constraints
have been obtained by \citex{GaspersSzeider11}.

\section{Bayesian Reasoning} 

\emph{Bayesian networks} (BNs) have emerged as a general representation
scheme for uncertain knowledge \cite{Pearl10}.  A BN models a set of
stochastic variables, the independencies among these variables, and a
joint probability distribution over these variables.  For simplicity we
consider the important special case where the stochastic variables are
Boolean. The variables and independencies are modelled in the BN by a
directed acyclic graph $G = (V,A)$, the joint probability distribution
is given by a table $T_v$ for each node $v \in V$ which defines a
probability $T_{v|U}$ for each possible instantiation
$U=(d_1,\dots,d_s)\in \{\True,\False\}^s$ of the parents
$v_1,\dots,v_s$ of $v$ in $G$. The probability $\Prob(U)$ of a complete
instantiation $U$ of the variables of~$G$ is given by the product of
$T_{v|U}$ over all variables~$v$.  We consider the problem
\textbf{Positive-BN-Inference} which takes as input a Boolean BN $(G,T)$
and a variable $v$, and asks whether $\Prob(v=\True) > 0$.  
\longversion{The problem
is $\NP$-complete \cite{Cooper90} and moves from NP to \#P if we ask to
compute $\Prob(v=\text{true})$ \cite{Roth96}.}
\shortversion{The problem
is $\NP$-complete \cite{Cooper90}.}
The problem can be solved
in polynomial time if the BN is \emph{singly connected}, i.e, if there
is at most one undirected path between any two variables
\cite{Pearl88}. It is natural to parametrize the problem by the number
of variables one must delete in order to make the BN singly connected
(the deleted variables form a \emph{loop cutset}). In fact,
$\PR{Positive-BN-Inference}{loop cutset size}$ is easily seen to be
fixed-parameter tractable as we can determine whether $\Prob(v=\True)>0$
by taking the maximum of $\Prob(v=\True \mid U)$ over all $2^k$ possible
instantiations of the $k$ cutset variables, each of which requires
processing of a singly connected network.  However, although
fixed-parameter tractable, it is unlikely that the problem admits a
polynomial kernel.
\begin{THE}
  $\PR{Positive-BN-Inference}{loop cutset size}$ does not admit a polynomial
  kernel unless the Polynomial Hierarchy collapses.
\end{THE} 
\begin{proof} (Sketch.)  We give a polynomial parameter transformation
  from $\PR{SAT}{vars}$ and apply Theorem~\ref{the:trans}.  The
  reduction is based on the reduction from 3SAT given by
  \citex{Cooper90}. However, we need to allow clauses with an arbitrary
  number of literals since, as observed above, $\PR{3SAT}{vars}$ has a
  polynomial kernel.  Let $F$ be a CNF formula on $n$ variables.  We
  construct a BN $(G,T)$ such that for a variable $v$ we have
  $\Prob(v=\True) > 0$ if and only if $F$ is satisfiable. Cooper uses
  \emph{input nodes} $u_i$ for representing variables of~$F$,
  \emph{clause nodes} $c_i$ for representing the clauses of $F$, and
  \emph{conjunction nodes} $d_i$ for representing the conjunction of the
  clauses.  We proceed similarly, however, we cannot represent a clause
  of large size with a single clause node $c_i$, as the required table
  $T_{c_i}$ would be of exponential size.
  \shortversion{ Therefore we split clauses
  containing more than 3 literals into several clause nodes. For
  instance, a clause node $c_1$ with parents $u_1,u_2,u_3$ is split into
  clause nodes $c_1,c_2$ where $c_1$ has parents $u_1,u_2$ and $c_2$ has
  parents $c_1,u_3$.}
\longversion{
  Therefore we split clauses
  containing more than 3 literals into several clause nodes, as
  indicated in 
 Figure~\ref{fig:BN}.
 \begin{figure}[tbhp]
   \centering
  \begin{tikzpicture}[xscale=1.4,yscale=0.7]
    \small
    \draw
    (0,0)   node (u1) {$u_1$}
    (1,0)   node (u2) {$u_2$}
    (2,0)   node (u3) {$u_3$}
    (3,0)   node (u4) {$u_4$}
    (.5,-.7)   node (c1) {$c_1$}
    (1.5,-1.2)   node (c2) {$c_2$}
    (2.5,-1.5)   node (c3) {$c_3$}
;
\draw[->] (u1)--(c1);
\draw[->] (u2)--(c1);
\draw[->] (c1)--(c2);
\draw[->] (u3)--(c2);
\draw[->] (c2)--(c3);
\draw[->] (u4)--(c3);
 \end{tikzpicture}%
\vspace{-4pt}
 \caption{BN representation of a clause on four
     literals.} \label{fig:BN}
 \end{figure}
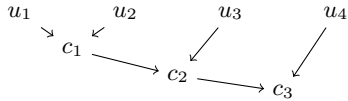}
It remains to observe that the set of input nodes $E=\{u_1,\dots,u_n\}$
is a loop cutset of the constructed BN, hence we have indeed  a
polynomial parameter transformation from $\PR{SAT}{vars}$  to
$\PR{Positive-BN-Inference}{loop cutset size}$. The result follows by
Theorem~\ref{the:trans}.
\end{proof}

\section{Nonmonotonic Reasoning}

\emph{Logic programming with negation} under the stable model semantics
is a well-studied form of nonmonotonic reasoning
\cite{GelfondLifschitz88,MarekTruszczynski99}.  A (normal) \emph{logic
  program}~$P$ is a finite set of rules $r$ of the form
\longversion{
  \[h \longleftarrow a_1 \wedge \dots \wedge a_m \wedge \neg b_1 \wedge
  \dots \wedge \neg b_n\] } \shortversion{$(h \leftarrow a_1 \wedge
  \dots \wedge a_m \wedge \neg b_1 \wedge \dots \wedge \neg b_n)$} where
$h,a_i,b_i$ are \emph{atoms}, where $h$ forms the head and the $a_i,b_i$
from the body of $r$.  We write $H(r)=h$, $B^+(r)=\{a_1,\dots,a_m\}$,
and $B^-(r)=\{b_1,\dots,b_n\}$. Let~$I$ be a finite set of atoms.  The
\emph{GF reduct} $P^I$ of a logic program~$P$ under $I$ is the program
obtained from~$P$ by removing all rules $r$ with $B^-(r)\cap I\neq
\emptyset$, and removing from the body of each remaining rule $r'$ all
literals $\neg b$ with $b\in I$.  $I$ is a \emph{stable model} of $P$ if
$I$ is a minimal model of $P^I$, i.e., if (i) for each rule $r\in P^I$
with $B^+(r)\subseteq I$ we have $H(r)\in I$, and (ii) there is no
proper subset of $I$ with this property. The \emph{undirected dependency
  graph} $U(P)$ of $P$ is formed as follows. We take the atoms of $P$ as
vertices and add an edge $x-y$ between two atoms $x,y$ if there is a
rule $r\in P$ with $H(r)=x$ and $y\in B^+(r)$, and we add a path $x-u-y$
if $H(r)=x$ and $y\in B^-(r)$ ($u$ is a new vertex of degree 2). The
\emph{feedback width} of $P$ is the size of a smallest set $V$ of atoms
such that every cycle of $U(P)$ runs through an atom in $V$.

A fundamental computational problems is \textbf{Stable Model Existence
  (SME)}, which asks whether a given normal logic program has a stable
model.  The problem is well-known to be
$\NP$-complete~\cite{MarekTruszczynski91}.
\citex{GottlobScarcelloSideri02} showed that $\PR{SME}{feedback width}$
is fixed-parameter tractable (see \citex{FichteSzeider11} for
generalizations).  We show that this result cannot be strengthened with
respect to a polynomial kernel.
\begin{THE}
  $\PR{SME}{feedback width}$ does not admit a polynomial kernel unless
  the Polynomial Hierarchy collapses.
\end{THE}
\shortversion{
  \begin{proof}(Sketch.)  \citex{Niemela99} describes a polynomial-time
    transformation that maps a CNF formula $F$ to a logic program $P$
    such that $F$ is satisfiable if and only if $P$ has a stable
    model. From the details of the construction it is easy to observe
    that the feedback width of $P$ is at most twice the number of
    variables in~$F$, hence we have a polynomial parameter
    transformation from $\PR{SAT}{vars}$ to $\PR{SME}{feedback
      width}$. The result follows by Theorem~\ref{the:trans}.
\end{proof}
} \longversion{\begin{proof}(Sketch.)  We give a polynomial parameter
    transformation from $\PR{SAT}{vars}$ to $\textbf{SME}($feedback width$)$
    using a construction of \citex{Niemela99}. Given a CNF formula
    $F$ on $n$ variables, we construct a logic program $P$ as
    follows. For each variable $x$ of $F$ we take two atoms $x$ and
    $\hat x$ and include the rules $(\hat x \leftarrow \neg x)$ and $(x
    \leftarrow \neg \hat x)$; for each clause $C$ of $F$ we take an atom
    $c$ and include for each positive literal $a$ of $C$ the rule $(c
    \leftarrow a)$, and for each negative literal $\neg a$ of $C$ the
    rule $(c \leftarrow \hat a)$; finally, we take two atoms $s$ and $f$
    and include the rule $(f \leftarrow \neg f \wedge \neg s)$ and for
    each clause $C$ of $F$ the rule $(s \leftarrow c)$.  $F$ is
    satisfiable if and only if $P$ has a stable model
    \cite{Niemela99}. It remains to observe that each cycle of $U(P)$
    runs through a vertex in $V=\SB x,\hat x \SM x \in
    \text{vars}(F)\SE$, hence the feedback width of $P$ is at most $2n$.
    Hence we have a polynomial parameter transformation from
    $\PR{SAT}{vars}$ to $\PR{SME}{feedback width}$. The result follows
    by Theorem~\ref{the:trans}.
\end{proof}
}

\shortversion{\vspace{-4pt}}
\section{Conclusion}
We have established super-polynomial kernel lower bounds for a wide
range of important AI problems, providing firm limitations for the power
of polynomial-time preprocessing for these problems. We conclude from
these results that in contrast to many optimization problems (see
Section 1), typical AI problems do not admit polynomial kernels.  Our
results suggest the consideration of alternative approaches.  For
example, it might still be possible that some of the considered problems
admit polynomially sized Turing kernels, i.e., a polynomial-time
preprocessing to a Boolean combination of a polynomial number of
polynomial kernels. In the area of optimization, parameterized problems
are known that do not admit polynomial kernels but admit polynomial
Turing kernels~\cite{FernauEal09}.
    This suggests a theoretical and empirical study of Turing kernels for
the AI problems considered.

\longversion{}

\shortversion{

}

\end{document}